\documentclass{article}

\usepackage{arxiv}

\usepackage[utf8]{inputenc} 
\usepackage[T1]{fontenc}    
\usepackage{hyperref}       
\usepackage{url}            
\usepackage{booktabs}       
\usepackage{amsfonts}       
\usepackage{nicefrac}       
\usepackage{microtype}      
\usepackage{lipsum}
\usepackage{textcomp}
\usepackage{graphicx}
\usepackage[
  backend=bibtex,
  style=numeric,
  sorting=none,
  natbib=true
]{biblatex}
\usepackage{amsmath}
\graphicspath{ {./fig/} }

\title{Learning-based Strategy for Composite Robot Assembly Skill Adaptation}

\author{
 Khalil Abuibaid \\
  Chair of Machine Tools and Control System,\\
  RPTU University Kaiserslautern-Landau,\\
  67663 Kaiserslautern,  Germany \\
  \texttt{khalil.abuibaid@rptu.de} \\
   \And
 Aleksandr Sidorenko \\
  Innovative Factory Systems,\\
  German Institute of Artificial Intelligence,\\
  67663 Kaiserslautern, Germany, \\
  \texttt{aleksandr.sidorenko@dfki.de} \\
  \And
 Achim Wagner \\
  Innovative Factory Systems,\\
  German Institute of Artificial Intelligence,\\
  67663 Kaiserslautern, Germany,\\
  \texttt{achim.wagner@dfki.de} \\
    \And
    Martin Ruskowski \\
Chair of Machine Tools and Control Systems, \\ 
RPTU University Kaiserslautern-Landau,  \\
67663 Kaiserslautern, Germany  \\
  and  \\
  Innovative Factory Systems,\\
  German Institute of Artificial Intelligence,\\
  67663 Kaiserslautern, Germany,\\
  \texttt{martin.ruskowski@dfki.de} \\
}

\begin{document}
\maketitle
\begin{abstract}
Contact-rich robotic skills remain challenging for industrial robots due to tight geometric tolerances, frictional variability, and uncertain contact dynamics, particularly when using position-controlled manipulators. This paper presents a reusable and encapsulated skill-based strategy for peg-in-hole assembly, in which adaptation is achieved through Residual Reinforcement Learning (RRL). The assembly process is represented using composite skills with explicit pre-, post-, and invariant conditions, enabling modularity, reusability, and well-defined execution semantics across task variations. Safety and sample efficiency are promoted through RRL by restricting adaptation to residual refinements within each skill during contact-rich interactions, while the overall skill structure and execution flow remain invariant. The proposed approach is evaluated in MuJoCo simulation on a UR5e robot equipped with a Robotiq gripper and trained using SAC and JAX. Results demonstrate that the proposed formulation enables robust execution of assembly skills, highlighting its suitability for industrial automation.
\end{abstract}
\section{Introduction}

Robotic assembly where precision mechanical components remains a longstanding challenge in industrial automation alone, particularly when the task involves low clearance and contact-rich interactions. Despite advances in compliant control, many of these assembly operations remain largely manual due to the sensitivity of the contact dynamics and the lack of generalizable autonomous solutions \cite{siciliano_springer_2008}.

Reinforcement Learning (RL) has shown strong potential for acquiring complex manipulation behaviors directly from interaction \cite{kalakrishnan2011learning,luo_deep_2018}. However, in contact-rich assembly scenarios, model-free RL methods suffer from sample inefficiency, unsafe exploration, and limited generalization, which significantly restrict their applicability in industrial environments. RRL addresses safety and sample efficiency by confining learning to residual refinements around a structured hybrid force–motion controller, thereby avoiding unstructured exploration in contact-rich assembly \cite{johannink_residual_2019,abuibaid_federated_2025,abuibaid_sustainable_2025,beltran-hernandez_learning_2020}.

In parallel, skill-based engineering approaches provide modularity, reusability, and explicit execution semantics for assembly tasks \cite{wagner2025skillbased,sidorenko_skills_2024}. By expressing domain expertise as predefined skills along with their starting and enabling conditions, such frameworks support dependable and consistently repeatable execution. Nevertheless, nominal skill-based methods rely on fixed parameters and handcrafted controllers, limiting their adaptability under varying contact conditions.

This work bridges these limitations by proposing a reusable and encapsulated skill-based assembly augmented with RRL that addresses contact conditions, including high friction, pose misalignment, and contact uncertainty.  Assembly, like any other manipulation function, is represented as a composition of alignment and insertion skills with explicit pre-, post-, and invariant conditions, while learning is strictly confined to residual refinements within each skill. The proposed approach is evaluated in simulation using a UR5e robot equipped with a Robotiq gripper. 

The remainder of this paper is organized as follows. Section~\ref{sec:relatedwork} reviews related work on skill-based and learning-based assembly strategies. Section~\ref{sec:method} presents the proposed encapsulated skill formulation and learning approach. Section~\ref{sec:expermental} reports the experimental evaluation, and Section~\ref{sec:conculsion} discusses the result and concludes the paper with directions for future research.

\section{Related Work} \label{sec:relatedwork}

Early studies demonstrated that deep RL can enable robotic skill acquisition for contact-rich assembly by learning directly from interaction, using contact-state representations and sparse success-based rewards to handle pose uncertainty in real industrial setups \cite{luo_deep_2018,li_robot_2019}. Subsequently, \cite{li_flexible_2022} used a digital-twin environment for safe DRL training and achieved high success rates in peg insertion. Assembly performance further benefits from structured perception and task decomposition, where combining visual and force modalities has been shown to enhance robustness to misalignment and shape variations \cite{song_skill_2021,ahn_robotic_2023}.

Beyond end-to-end learning, classical assembly frameworks remain foundational. Taxonomies of manipulation primitives provide a structured decomposition of fine assembly skills \cite{suarez-ruiz_framework_2016}. Extending this line of work, \cite{li_manipulation_2019} employed off-policy deep RL with joint-torque safety constraints to learn manipulation skills directly on a real robot without prior task knowledge. To improve data efficiency and reduce manual controller tuning, \cite{lammle_skill-based_2020} introduced a skill-based RL framework that learns force and position parameters of pre-defined assembly skills in simulation rather than learning low-level control policies from scratch. More recently, \cite{stranghoner_share-rl_2025} proposed SHaRe-RL, a structured and interactive RL framework that combines skill decomposition, human demonstrations, and constraints in the compliance to enable safe and sample-efficient online learning for contact-rich industrial assembly tasks. 

Despite significant progress, limitations remain. Model-free RL methods are sample-inefficient and difficult to deploy safely in contact-rich assembly due to unstructured exploration \cite{luo_deep_2018,li_flexible_2022}. Nominal skill-based frameworks enable safe and reusable execution through predefined force–motion primitives but offer limited adaptability due to fixed parameters \cite{wagner2025skillbased,sidorenko_skills_2024}. Learning-based skill refinement improves data efficiency within structured skills \cite{li_manipulation_2019,lammle_skill-based_2020,stranghoner_share-rl_2025}, yet often relies on external supervision or interaction-specific adaptation mechanisms, limiting full encapsulation and autonomous reuse of assembly skills. Consequently, achieving adaptive, yet reusable assembly skills through fully self-contained residual learning remains an open challenge.

\section{Methodology} \label{sec:method}
\subsection{Structured Strategy for Assembly}
Following principles of compliant motion \cite{siciliano_springer_2008} and skill-based robotic frameworks \cite{wagner2025skillbased,sidorenko_skills_2024}, the assembly process is decomposed into two composite interaction skills, namely, an alignment and an insertion, where the skill is formulated as follows
\begin{equation}
            \mathbb{S}  = \{C_{pre},C_{inv},C_{post},F_{Skill}\},
    \label{eq:skill_definition}
\end{equation} 
where $C_{pre}$ and $C_{post}$ are conditions that represent states of the world when a skill can be started and after its successful execution, respectively. While, $C_{inv}$ defines conditions that must hold true during the process, and $F_{Skill}$ is the function that help transiting during the assembly procedure \cite{wagner2025skillbased,sidorenko_skills_2024}.  The skill-based approach aligns with current trends in learning contact-interaction behaviors \cite{stranghoner_share-rl_2025} and with learning-based peg-in-hole strategies that combine force control with RL \cite{zou_learning-based_2020,tu_robotic_2024}.  

\subsubsection{Alignment Skill}  
This skill addresses the initial radial, angular, and axial misalignment between the peg and the hole by performing compliant probing actions and interpreting instantaneous force and torque feedback to determine misalignment direction and magnitude, which enables the system to explore and localize the bore entrance.  Note that, there are two parameters, i.e., thresholds, which define the skills, namely the positional goal and force goal.

\subsubsection{Insertion Skill}  
After alignment is achieved, the robot transitions to an insertion skill in which motion along the insertion direction is regulated through hybrid force-motion controller. The insertion progress is modulated according to measured interaction forces to avoid jamming, surface scratching, or excessive contact loads.  Recent learning-based approaches \cite{zou_learning-based_2020,tu_robotic_2024,beltran-hernandez_learning_2020} demonstrate that combining analytical force-control schemes with RRL improves insertion robustness under geometric uncertainty.

\begin{figure}[h!]
    \centering
    \includegraphics[width=0.950\linewidth]{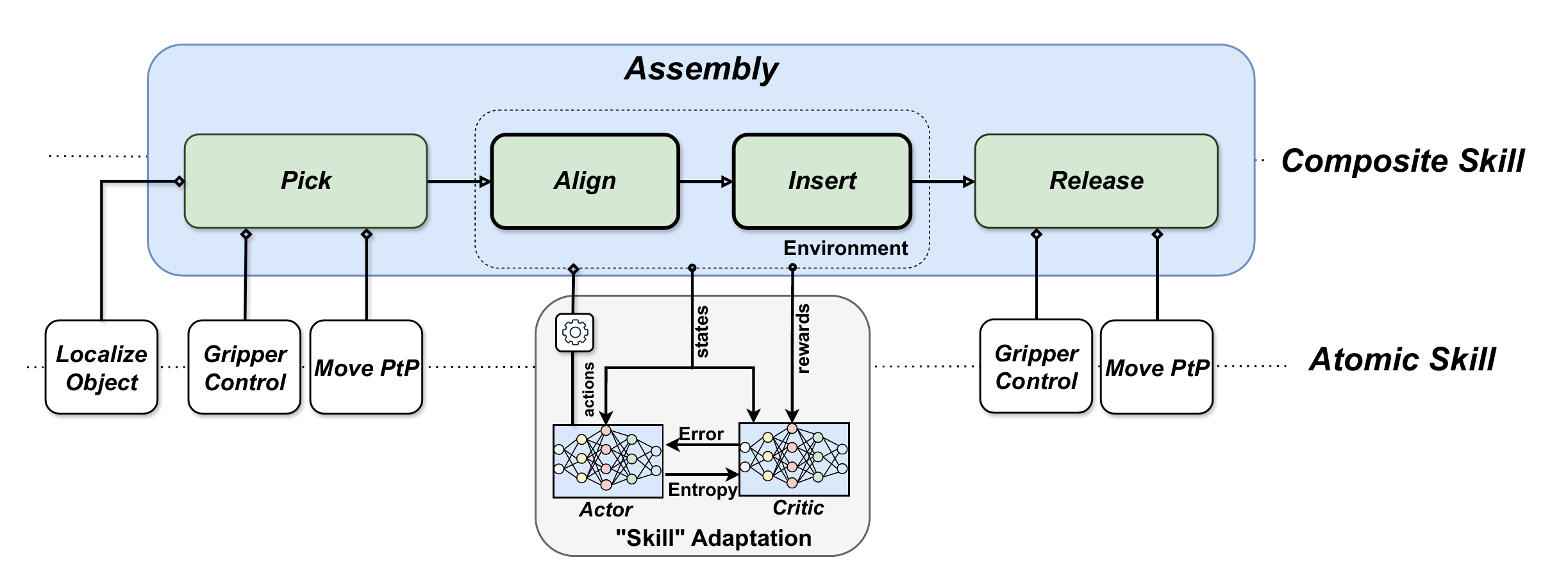}
    \caption{Encapsulated skill hierarchy for the assembly task.}
    \label{fig:skill_based}
\end{figure}

As illustrated in Fig.~\ref{fig:skill_based}, the hierarchical decomposition yields modular and reusable assembly skills with explicit execution semantics. Composite skills define a flexible task structure depending on actual conditions, while atomic skills preserve fixed execution behavior, enabling adaptation without altering the orchestration sequence and supporting reuse across different robots.

\subsection{Residual Reinforcement Learning} \label{subsec:hybrid}

The execution of the assembly requires regulating both motion and contact forces in cartesian space. We employ a hybrid force-motion controller~\cite{siciliano_springer_2008} and augment it with RRL~\cite{johannink_residual_2019,abuibaid_sustainable_2025,abuibaid_federated_2025}.

\subsubsection{Hybrid Force-Motion Controller}
The nominal hybrid controller computes with motion and force subsystems
\begin{equation}
    u_{\mathrm{nom}}(t) = (I - \Sigma)\!\left(K_{p}^{X} X_{e} + K_{d}^{X} \dot{X}\right) + (\Sigma)\left(K_{p}^{F} F_{e}
    + K_{i}^{F} \int_{0}^{t} F_{e}(\tau)\, d\tau \right),
\end{equation}
where $\Sigma = \mathrm{diag}(\sigma_1,\ldots,\sigma_6)$ selects motion or force to control on each axis.  
The gains $K_{d}^{X}$ and $K_{i}^{F}$ ensure damping and integral compensation during contact, where they are defined as $K_{d}^{X}  =2\times(K_p^X)^{\frac{1}{2}}$ and $K_{i}^{F} = 0.01\times K_p^F$.

\subsubsection{Learning-Based Algorithm}
To enhance robustness during alignment and insertion skills, using RRL helps adapting the uncertainties~\cite{beltran-hernandez_learning_2020}. The executed command becomes
\begin{equation}
    u(t) = u_{\mathrm{nom}}(t) + u_{\mathrm{rl}}(t),
\end{equation}
which modulates the nominal controller and a residual action $u_{\mathrm{rl}}$ that refines cartesian space behavior near contact by generating the optimal motion trajectory, $X_{a}$. The policy's action, i.e., $A_t$ outputs $[\, X_{a},\ K_{p}^{X},\ K_{p}^{F},\ \Sigma \,]_t^T,$
while the policy receives a temporal stack of cartesian space features $S \in \mathbb{R}^{18 \times l} $ in  $S_t = \big[X_e, \; \dot{X},\;F_e\;\big]_t^T$, which provides a 6D error pose and velocity representation and a 6D error force/torque, consistent with the assembly-state definition in~\cite{song_skill_2021}. Hence, $X_e = (x_e, y_e, z_e, \alpha_e, \beta_e, \gamma_e)$, $\dot{X}  = (\dot{x},\dot{y},\dot{z},\dot{\alpha},\dot{\beta},\dot{\gamma} ) $ and $F_e  = (f_e^x, f_e^y, f_e^z, \tau_e^x, \tau_e^y, \tau_e^z)$. The rewards function is formulated as follows; 
\begin{equation}
r_{sparse} (t) = 500 \cdot 1(\text{insert}) + 50 \cdot 1(\text{align}) - 80 \cdot 1(\text{collision})
\end{equation}
The dense reward is defined as $r_{dense}(t) = \mathbf{w}^T \|\mathbf{S}_t\|_2$, where $\|\mathbf{S}_t\|_2$ is the vector of component-wise $L_2$ norms and $\mathbf{w} = [-5, -0.001, -0.005]^T$.
\begin{figure}[h]
    \centering
    \includegraphics[width=0.950\linewidth]{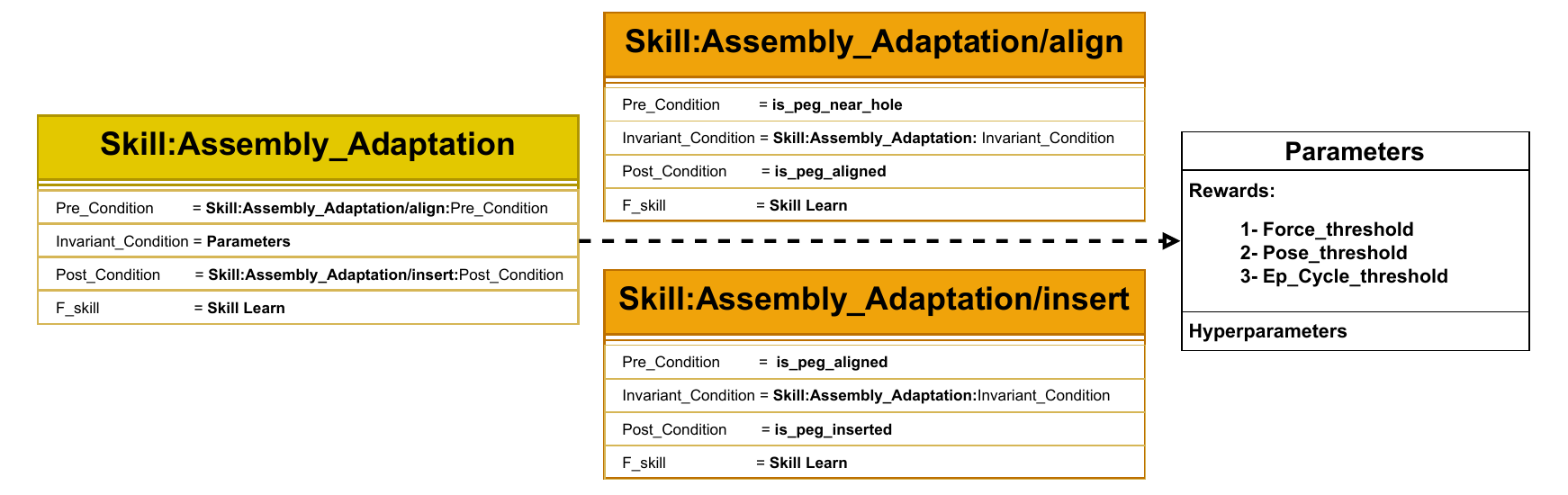}
    \caption{Paradigm of Skill-based programming for assembly skill adaption by describing \eqref{eq:skill_definition}.}
    \label{fig:skill_based_assembly}
\end{figure}

Since the robot control frequency runs at $500~Hz$ and the RRL frequency is $40~Hz$, it is recommended to store the history length $l$ of $S$ in a buffer for training the policy of each loop and use it \cite{stable-baselines3}. However, to train the RRL to do the assembly skill adaptation, the problem must be divided into smaller training skills, namely aligning and insertion skills, see Figure \ref{fig:skill_based_assembly}. The $C_{post}$ of align and insert skills are defined as $x_{g}^{align}$ and $x_{g}^{insert}$.

\begin{figure}[h]
    \centering
    \includegraphics[width=0.950\linewidth]{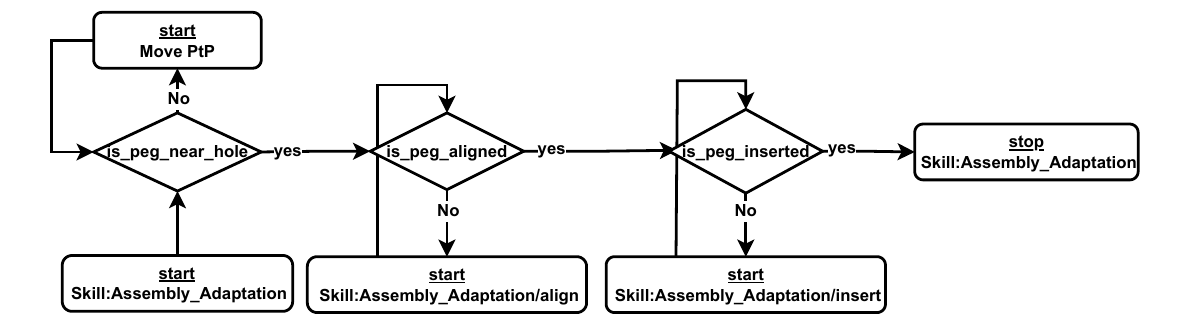}
    \caption{Activity Diagram of the Assembly Skill adaptation program.}
    \label{fig:skill_based_assembly}
\end{figure}

The assembly skill adaptation is treated as skill composed of composite alignment and insertion skill, where $F_{\text{skill}}$ is using RRL, as illustrated in Fig.~\ref{fig:skill_based_assembly}. The post-condition of alignment, i.e., $x_{g}^{align}$ corresponds to the pre-condition of the insert skill. While The post-condition of the insert corresponds to the final insertion goal $x_{g}^{insert}$. If the assembly adapatation is interrupted during the insertion phase, the execution is resumed directly from the insertion skill without re-executing the alignment skill, thereby avoiding unnecessary repetition and improving execution efficiency. During execution, $F_{\text{skill}}$ ensures a smooth transition between these conditions through a blending function,
\begin{equation}
    x_{g}^{insert}(t)
    = (1-\alpha_{\mathrm{smooth}})\,x_{g}^{align}
    + \alpha_{\mathrm{smooth}}\,x_{g}^{insert},
\end{equation}
where $\alpha_{\mathrm{smooth}} \in [0,1]$. This formulation ensures continuous execution of the assembly skill by smoothly transitioning from the alignment phase to the insertion phase, while allowing the controller to respond to evolving contact force profiles. The $C_{pre}$ of alignment skill is initialized with randomized pose and inclination of peg, with $\alpha,\beta \in [0.1, 0.15]~\text{rad}$, whereas the $C_{post}$ of the insert skill is obtained assumingly through visual detection. 

\section{Simulation Experimental Setup} \label{sec:expermental}
\begin{figure}[h]
    \centering
    \includegraphics[width=1.0\linewidth]{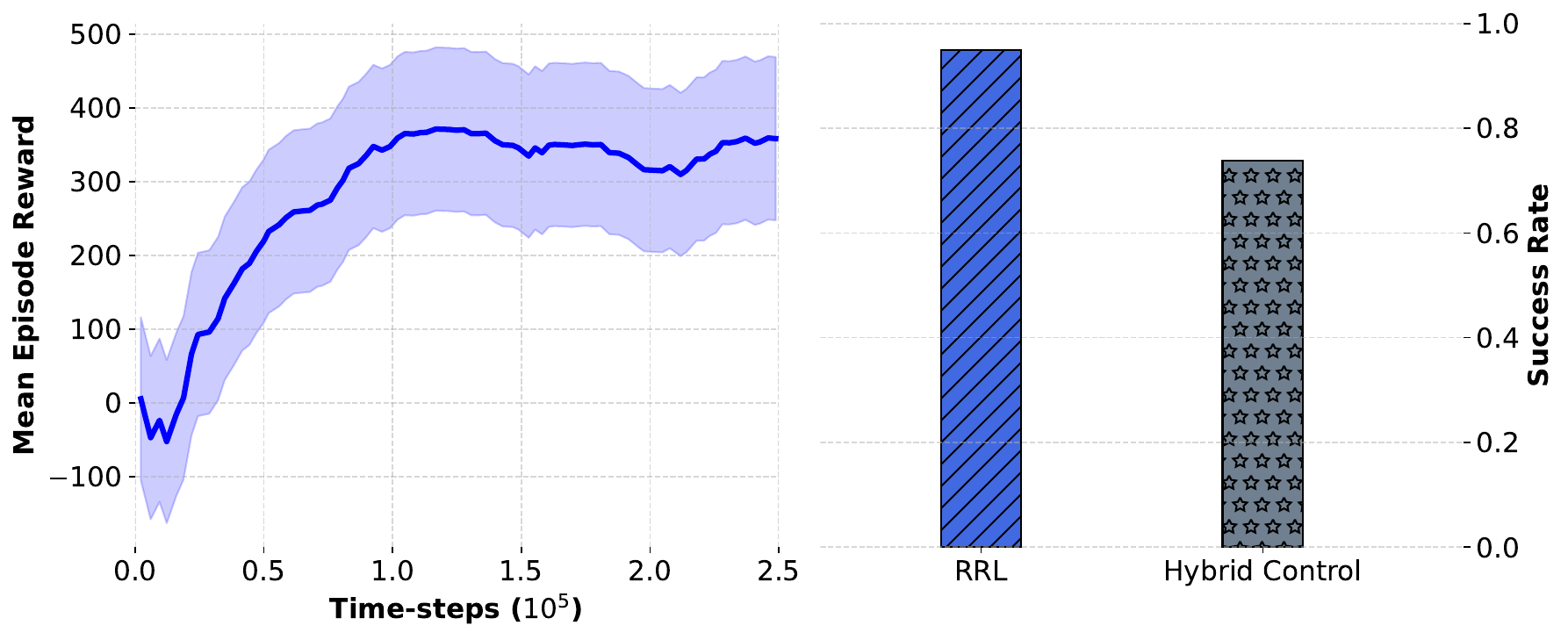}
    \caption{Result of the assembly skill adaptation training and testing.}
    \label{fig:training}
\end{figure}

The evaluation is conducted in MuJoCo~\cite{mujoco} on a UR5e robot equipped with a Robotiq 2F-85 gripper. The Skill Learn is implemented using the sb3 with JAX (sbx) framework~\cite{stable-baselines3}. The actor and critic networks are [400,400,200]. A learning rate of $1\times10^{-4}$, batch size of $400$, and replay buffer size of $1\times10^{6}$ are used for the experiments.
As illustrated in Figure \ref{fig:training}, the training is performed for $2.5\times10^{5}$ simulation steps at an effective control frequency of approximately $13~Hz$. To reflect realistic contact conditions, a square peg geometry is used, and friction coefficients for both peg and hole surfaces are defined up to $0.2$ during training. Hence, the best model is saved for testing. For a fair assessment, the proposed RRL approach is compared against a nominal hybrid force–motion controller in the context of assembly skill adaptation, where RRL demonstrates superior performance relative to the nominal baseline as shown in Figure \ref{fig:training}.

To promote reusability, generalization, and robustness, noise is injected into joint states and wrench measurements during training, preventing overfitting and enabling reuse of the learned skill across varying systems. Additionally, the exploration–exploitation mechanism of SAC is leveraged to further improve policy robustness.

\section{Conclusion and Discussion} \label{sec:conculsion}
This work presents a reusable and encapsulated skill-based strategy for contact-rich peg-in-hole assembly, in which adaptation is realized through RRL. The assembly skill adaptation is modeled as a composition of alignment and insertion skills with invariant execution semantics, confining learning to residual refinements within each skill. This design promotes safety and sample efficiency while preserving the modular structure of the nominal controller.

From a broader perspective, the proposed formulation demonstrates that robust adaptation to pose misalignment and tight geometric tolerances under high-friction contact conditions can be achieved without manual tuning skill structure. This separation between skill execution and learning enables reuse of the same skill abstraction across varying assembly conditions, which is a key requirement for industrial deployment. Future work will extend the framework toward learning multiple specialized residual models per skill, rather than relying on a single centralized model for both alignment and insertion, and will further investigate transfer to real-world robotic systems.

\section*{Acknowledgement}
This work has received funding from the European Union’s Horizon Europe research and innovation programme under grant agreement No. 101138782 (RAASCEMAN).
%
%
\printbibliography

\end{document}